\documentclass[conference]{IEEEtran}
\IEEEoverridecommandlockouts
\usepackage{cite}

\usepackage[numbers,sort&compress]{natbib}
\usepackage[utf8]{inputenc} 
\usepackage[T1]{fontenc}    
\usepackage{hyperref}       
\usepackage{url}            
\usepackage{booktabs}       
\usepackage{mathtools,amssymb}
\usepackage{amsfonts}       
\usepackage{nicefrac}       
\usepackage{microtype}      
\usepackage{pgfplots,pgfplotstable}
\pgfplotsset{compat=1.14}
\usepackage{array,colortbl}
\usepackage{xcolor}
\usepackage{algorithm,algorithmicx,algpseudocode}
\usepackage[capitalise]{cleveref}
\usepackage{caption}
\usepackage{graphbox}
\usepackage{placeins}
\usepackage{wrapfig}
\usepackage{etoolbox}
\usepackage{subfloat}
\usepackage{graphicx}
\usepackage[thicklines]{cancel}

\newcommand{\bI}{\mathbf{I}}

\newcommand{\bzero}{\mathbf{0}}

\newcommand{\bx}{\mathbf{x}}
\newcommand{\by}{\mathbf{y}}
\newcommand{\bz}{\mathbf{z}}

\newcommand{\bepsilon}{{\boldsymbol{\epsilon}}}

\usepackage{amsmath}
\usepackage{caption}
\usepackage{subcaption}
\usepackage{bm}

\def\BibTeX{{\rm B\kern-.05em{\sc i\kern-.025em b}\kern-.08em
    T\kern-.1667em\lower.7ex\hbox{E}\kern-.125emX}}
\begin{document}

\title{
INDigo: An INN-Guided Probabilistic Diffusion Algorithm for Inverse Problems
\\
}

\author{\IEEEauthorblockN{Di You}
\IEEEauthorblockA{\textit{EEE Department} \\
\textit{ Imperial College London}\\
London, UK \\
di.you22@imperial.ac.uk}
\and
\IEEEauthorblockN{Andreas Floros}
\IEEEauthorblockA{\textit{EEE Department} \\
\textit{Imperial College London}\\
London, UK \\
andreas.floros18@imperial.ac.uk}
\and
\IEEEauthorblockN{Pier Luigi Dragotti}
\IEEEauthorblockA{\textit{EEE Department} \\
\textit{Imperial College London}\\
London, UK \\
p.dragotti@imperial.ac.uk}
}

\maketitle

\begin{abstract}
Recently it has been shown that using diffusion models for inverse problems can lead to remarkable results. However, these approaches require a closed-form expression of the degradation model and can not support complex degradations. To overcome this limitation, we propose a method (INDigo) that combines invertible neural networks (INN) and diffusion models for general inverse problems. Specifically, we train the forward process of INN to simulate an arbitrary degradation process and use the inverse as a reconstruction process. During the diffusion sampling process, we impose an additional data-consistency step that minimizes the distance between the intermediate result and the INN-optimized result at every iteration, where the INN-optimized image is composed of the coarse information given by the observed degraded image and the details generated by the diffusion process. With the help of INN, our algorithm effectively estimates the details lost in the degradation process and is no longer limited by the requirement of knowing the closed-form expression of the degradation model. Experiments demonstrate that our algorithm obtains competitive results compared with recently leading methods both quantitatively and visually. Moreover, our algorithm performs well on more complex degradation models and real-world low-quality images.

\end{abstract}

\begin{IEEEkeywords}
inverse problems, diffusion models, invertible neural networks.
\end{IEEEkeywords}

\section{Introduction}
In this paper, we focus on the problem of reconstructing a high-quality image $\mathbf{x}$ from noisy and degraded measurements $\mathbf{y}$. This inverse problem is typically modelled as follows:
\begin{align}
    \mathbf{y}& =\mathcal{H}(\mathbf{x})+\mathbf{n}
\end{align}
where $\mathcal{H}$ models the degradation process and $\mathbf{n} \sim\mathcal{N}(\mathbf{0},{\sigma_0}^2\mathbf{I})$ is additive noise. In this paper, we assume that $\mathcal{H}(\cdot)$ can be either linear or non-linear.
Many imaging tasks fall under this model including deblurring, super-resolution and removal of compression artefacts.

With the emergence of deep learning techniques, many deep learning-based algorithms for inverse problems have achieved excellent success and we refer to \cite{review} for a recent overview.

Recently, the generative prior of diffusion models \cite{ddpm,ddim,dhariwal2021diffusion,song2019generative,songscore} has become one of the most popular priors due to their remarkable ability to approximate the natural image manifold. {{A line of work \cite{ilvr,ccdf,Repaint,kadkhodaie2020solving,snips,DDRM,DDNM,pseudoinverse,dps,dolce} has focused on leveraging the learned score function as a generative prior of the data distribution to solve general inverse problems.
Among them, earlier works \cite{ilvr,ccdf,Repaint,kadkhodaie2020solving} have demonstrated the great ability of diffusion models for inverse problems. While promising, these methods are limited by the assumption that $\mathcal{H}(\cdot)$ is a linear operator and that there is no noise.
A classic example is ILVR, which leverages a trained diffusion model guided by the low-frequency information from a conditional image through the diffusion process.
}}

\begin{figure}[t]
    \centering
    \includegraphics[width=0.45\textwidth]{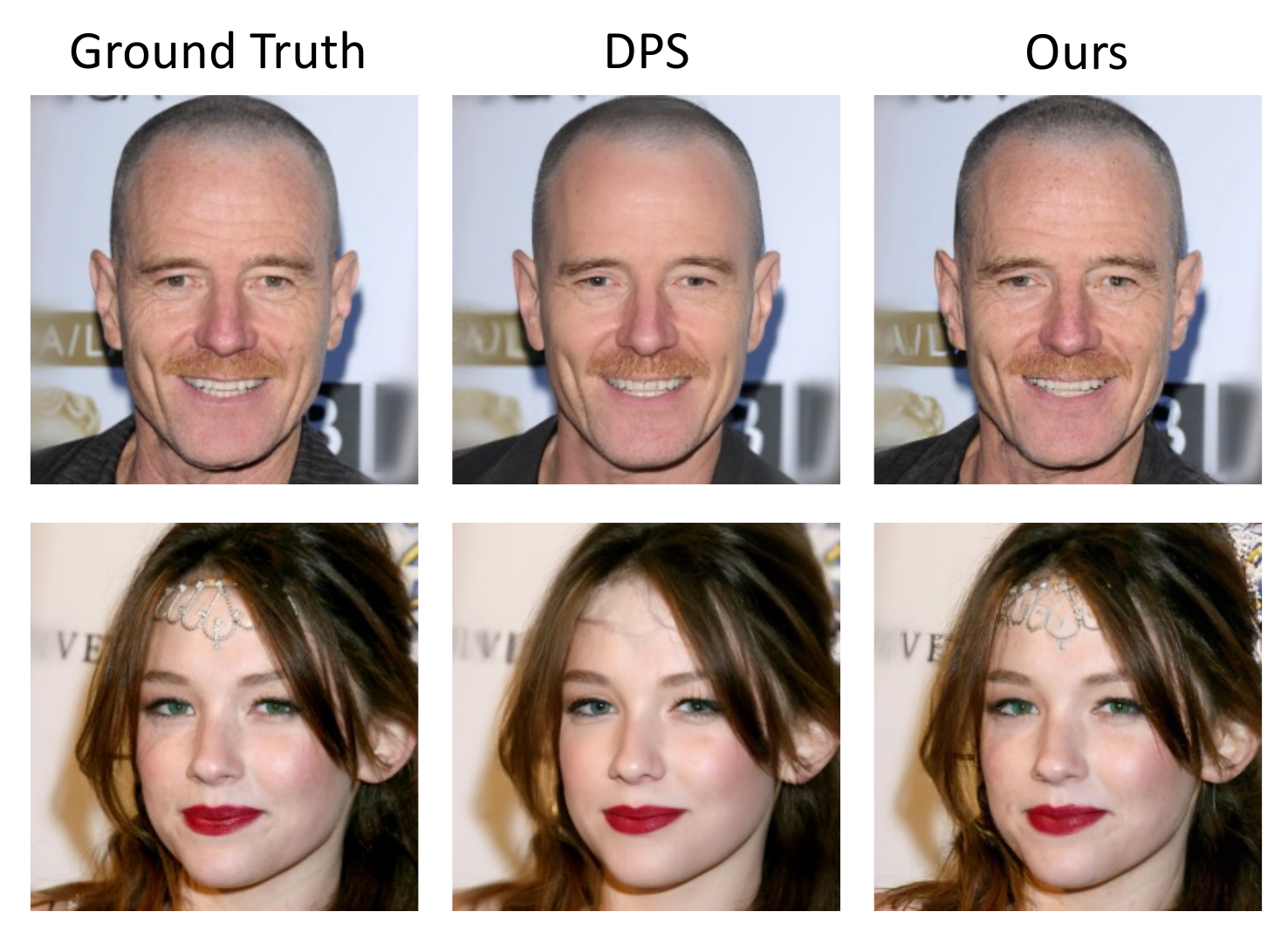}
\caption{Visual comparisons of DPS \cite{dps} and our method on solving super-resolution problem (x4) on CelebA HQ dataset.}
        \vspace{-0.5cm}
    \label{fig:results_linear}
\end{figure} 

Several approaches \cite{snips,DDRM,DDNM} have been proposed to solve noisy inverse problems using diffusion models. These methods run SVD/Range-Null space decomposition on intermediate results during iterations. Although their results on noisy cases show impressive reconstruction performance and good interpretability, they cannot solve non-linear inverse problems, e.g. Jpeg compression. Furthermore, they suffer from high computational complexity when handling more complicated degradation models.

More recently, several gradient-based methods \cite{pseudoinverse,dps,dolce} 
have been proposed to further generalize to non-linear noisy inverse problems.
During the $T$-step diffusion process, these methods impose an additional data-consistency step in the measurement domain. Specifically, at each iteration $t$ after sampling, they minimize the distance between $\mathbf{y}$ and $ \mathcal{H}(\mathbf{x}_{0,t})$, where $\mathbf{x}_{0,t}$ is the intermediate result of sampling.
This strategy avoids high-complexity computation and allows their models to have the potential to solve more complex inverse problems.
However, there are two main limitations of this type of approaches.
{Firstly}, imposing consistency in the measurement domain can blur the details generated by the diffusion process (see Fig.~\ref{fig:results_linear}). {Secondly}, all these methods require a closed-form expression of the degradation model $ \mathcal{H}(\cdot)$. However, the image processing pipeline of many modern imaging systems is so complex that it is often impossible to describe it with a closed-form expression.

To address the above issues, we propose a novel method that combines invertible neural networks (INN) with diffusion generative models. We introduce a Wavelet-inspired INN (WINN) where its forward part (WINN$_F$) splits the input image $\bx$ into a coarse version $\mathbf{c}$ and "details" $\mathbf{d}$. Given training data pairs $\left \{ \bx^{i}, \by^{i}\right \}_{i=1}^{N}$, WINN$_F$ is trained to simulate the degradation process {\color{black}{so that $\mathbf{c}$ is similar to $\by$}}, while $\mathbf{d}$ represents the lost details. At testing, given the degraded image $\by$, the diffusion process is conditioned using WINN. Specifically, at each step $t$, in the diffusion process,
instead of minimizing the distance between $\by$ and $\mathcal{H}(\bx_{0,t})$ as in \cite{pseudoinverse,dps,dolce}, we feed $\bx_{0,t}$ to WINN$_F$ to get its projection in measurement domain $\mathbf{c}$ and details $\mathbf{d}$. By replacing $\mathbf{c}$ with measurement $\by$, the INN-optimized $\hat\bx_{0,t}$ is generated by the {\color{black}{inverse}} process {\color{black}{WINN$_I$}}. Thus, the $\hat\bx_{0,t}$ is composed of {\color{black}{the coarse information given by $\by$}} and the details generated by the diffusion process. We then use $\hat\bx_{0,t}$ to update $\bx_{0,t}$ before moving to step $t-1$.

Therefore, we summarize the benefits of our algorithm as follows: \textbf{1)} Compared with existing gradient-based methods \cite{pseudoinverse,dps,dolce}, our design  estimates the details lost in the degradation process more effectively
in the data-consistency step. Furthermore, our solution is based on a more general assumption without the requirement of the closed-form expression of the degradation model, as we simulate it with INNs. 
\textbf{2)} Compared with methods based on SVD/Range-Null space decomposition \cite{snips,DDRM,DDNM}, our method avoid high-complexity computation and has the ability to solve more complex inverse problems.
\textbf{3)} Different from training a conditional diffusion model for a specific inverse problem \cite{saharia2022image,li2022srdiff,rombach2022high}, we train our INNs via only MSE loss and only modify the inference procedure to enable sampling from a conditional distribution. 
This strategy can make the diffusion model serve as a much stronger prior and lift the burden of simultaneous learning both data consistency and detail generation.
\textbf{4)} Extensive experiments show that our proposed model achieves state-of-the-art results compared with other recent leading methods in variety of inverse problems including super-resolution, jpeg compression and real degraded image reconstruction. 
\textbf{5)} To the best of our knowledge, this is the first attempt to combine the merits of the perfect reconstruction property of INN with strong generative prior of diffusion models for general inverse problems.

\begin{figure}
     \centering
     \begin{subfigure}[b]{0.45\textwidth}
         \centering
         \includegraphics[width=\textwidth]{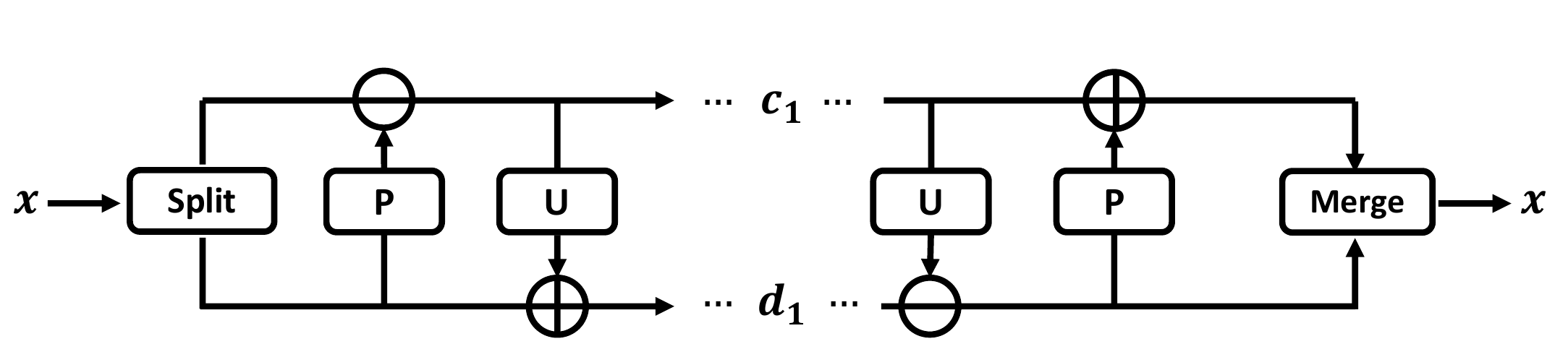}
         \caption{1-level lifting scheme}
         \label{fig:y equals x}
     \end{subfigure}
     \hfill
     \begin{subfigure}[b]{0.48\textwidth}
         \centering
         \includegraphics[width=\textwidth]{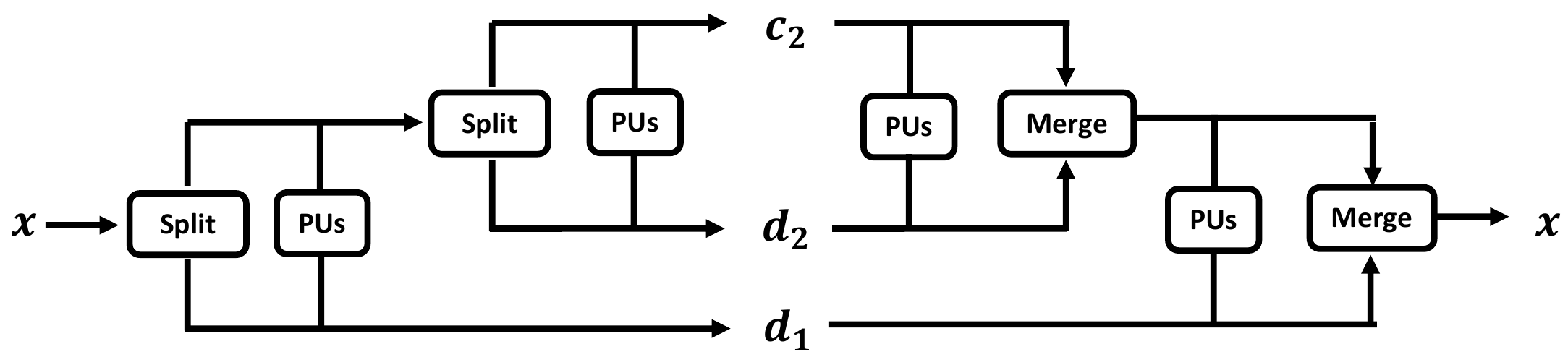}
         \caption{2-level lifting scheme}
         \label{fig:three sin x}
     \end{subfigure}
        \caption{The wavelet transform obtained using the lifting scheme.}
        \label{fig:lifting}
\end{figure}
\begin{figure*}[t]
    \centering
    \includegraphics[width=0.95\textwidth]{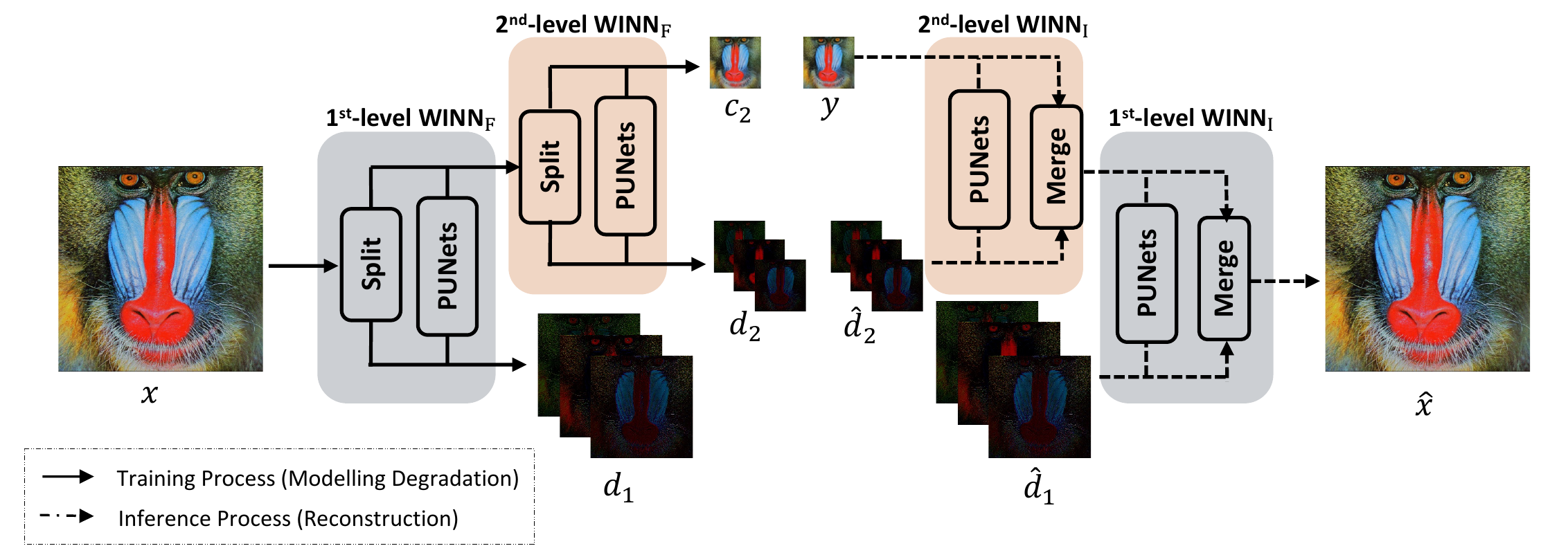}
\caption{ Overview of our wavelet-inspired invertible neural network (WINN) with 2 levels. We first train the forward process of WINN to guide the network to decompose the input image $\bx$ into measurements and lost information.
Then the inverse process of WINN can produce the reconstructed image $\hat\bx$ given measurement $\by$ and detail $\mathbf{\hat{d}}$ generated by the diffusion model.}
    \label{fig:framework}
\end{figure*}
\section{Background}
\label{gen_inst}

\subsection{Review the Diffusion Models}
Diffusion models, e.g., \cite{ddpm,ddim} sequentially corrupt training data with slowly increasing noise, and then learn to reverse this corruption in order to form a generative model of the data. Here we describe a classic diffusion model: denoising diffusion probabilistic models (DDPM)~\cite{ddpm}.
DDPM defines a $T$-step forward process transforming complex data distribution into simple Gaussian noise distribution and a $T$-step reverse process recovering data from noise. 
The forward process slowly adds random noise to data, where the added noise has a Gaussian distribution in the typical setting.
Consequently, the forward process yields the present state $\mathbf{x}_{t}$ from the previous state $\mathbf{x}_{t-1}$:
\begin{equation}
    q(\mathbf{x}_{t}|\mathbf{x}_{t-1})=\mathcal{N}(\mathbf{x}_{t};\sqrt{1-\beta_{t}}\mathbf{x}_{t-1},\beta_{t}\mathbf{I})\quad
    \label{eq:ddpm forward 1}
\end{equation}
where $\mathbf{x}_{t}$ is the noisy image at time-step $t$, $\beta_{t}$ is a predefined scale factor. As noted in \cite{ddpm}, the above process allows us to sample an arbitrary state $\mathbf{x}_{t}$ directly from the input $\mathbf{x}_{0}$ as follows:
\begin{equation}
\mathbf{x}_{t}=\sqrt{\Bar{\alpha}_{t}}\mathbf{x}_{0}+ \sqrt{1-\Bar{\alpha}_{t}} \boldsymbol{\epsilon} 
    \label{eq:xt}
\end{equation}
where $\alpha_{t} = 1- \beta_{t}$, $\quad \Bar{\alpha}_{t} = \prod_{i=0}^{t}\alpha_{i}$ and $\boldsymbol{\epsilon}\sim \mathcal{N}(0,\mathbf{I})$.
For the reverse process, we can calculate the posterior distribution $q(\mathbf{x}_{t-1}|\mathbf{x}_{t},\mathbf{x}_{0})$ using Bayes theorem and write the expression of $\mathbf{x}_{t-1}$ using Eq.~\ref{eq:xt} as follows:
\begin{equation}
\begin{split}
   \mathbf{x}_{t-1} = \frac{1}{\sqrt{\alpha_{t}}}	\left( \mathbf{x}_{t} - \frac{1-\alpha_{t}}{\sqrt{1-\Bar{\alpha}_{t}}} \boldsymbol{\epsilon} \right) + \sigma_t\mathbf{z}.
    \label{eq:ddpm reverse 1}
\end{split}
\end{equation}
where $\mathbf{\sigma}_{t}$$=$$\sqrt{\frac{1-\Bar{\alpha}_{t-1}}{1-\Bar{\alpha}_{t}}\beta_{t}}$ and $\mathbf{z}\sim \mathcal{N}(0,\mathbf{I})$.
To predict the noise $\boldsymbol{\epsilon}$ in the above equation, DDPM uses a neural network  $\boldsymbol{\epsilon}_{\boldsymbol{\theta}}(\mathbf{x}_{t},t)$ for each time-step $t$.
To train  $\boldsymbol{\epsilon}_{\boldsymbol{\theta}}(\mathbf{x}_{t},t)$, DDPM {\color{black}{uniformly samples a $t$ from $ \{1,...,T\}$}} and updates the network parameters $\boldsymbol{\theta}$ with the following gradient descent step:
\begin{equation}
    \nabla_{\boldsymbol{\theta}}||\boldsymbol{\epsilon}-\boldsymbol{\epsilon}_{\boldsymbol{\theta}}(\sqrt{\Bar{\alpha}_{t}}\mathbf{x}_{0} + \sqrt{1-\Bar{\alpha}_{t}}\boldsymbol{\epsilon},t)||^{2}_{2},
    \label{eq:7}
\end{equation}
where $\mathbf{x}_{0}$ is a clean image from the dataset and $\boldsymbol{\epsilon}\sim \mathcal{N}(0,\mathbf{I})$ is random noise. 
By replacing $\boldsymbol{\epsilon}$ with the approximator $\boldsymbol{\epsilon}_{\boldsymbol{\theta}}(\mathbf{x}_{t},t)$ in Eq.~\ref{eq:ddpm reverse 1} and iterating it T times,  DDPM can yield clean images $\mathbf{x}_{0}\sim q(\mathbf{x})$ from initial random noises $\mathbf{x}_{T}\sim\mathcal{N}(\mathbf{0},\mathbf{I})$, where $q(\mathbf{x})$ represents the image distribution in the training dataset.

\subsection{Invertible Neural Network based on Lifting scheme}
\label{background_inn}
The wavelet transform is widely used in many imaging applications because it is able to concentrate image features in a few large-magnitude wavelet coefficients, while small-value wavelet coefficients typically contain noise and can be shrunk or removed without affecting the image quality.
The lifting scheme \cite{daubechies1998factoring} is one of the options to construct a wavelet transform. As shown in Fig.~\ref{fig:lifting}(a), the lifting scheme first splits the signal into an even and an odd part. A predictor is used to predict the odd signal from the even part, therefore leading to a sparse residual signal. The update step is used to adjust the even signal based on the prediction error of the odd part to make it a better coarse version of the original signal. There can be multiple levels and multiple pairs of predictor and updater (see Fig.~\ref{fig:lifting}(b)). The lifting scheme can represent a transform with perfect reconstruction condition and any intermediate representation can be used to infer the input signal and the final representation.

Inspired by the above idea, Huang et al. \cite{winnet} propose to learn a non-linear wavelet-like transform with perfect reconstruction property using invertible neural networks with a structure inspired by the lifting scheme (LINN). 
Similar to the lifting scheme which splits the input signal and then alternates prediction and update \cite{, daubechies1998factoring},
LINN consists of splitting/merging operators and learnable predict and update networks. The design of predict and update networks will not 
affect the invertibility of LINN.

\begin{figure}
     \centering
     \vspace{-0.5cm}
     \begin{subfigure}[b]{0.4\textwidth}
         \centering
         \includegraphics[width=0.9\textwidth]{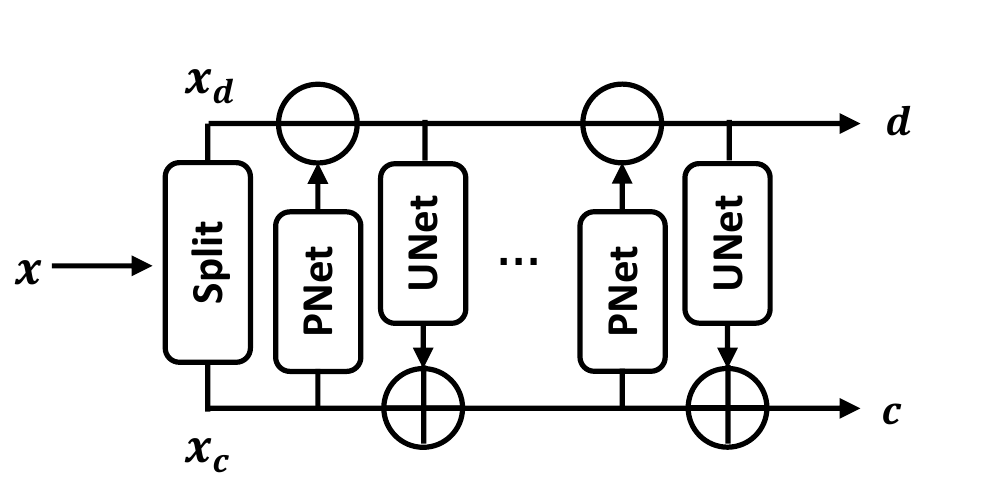}
                   \vspace{-0.4cm}
         \caption{}
         \label{fig:y equals x}
     \end{subfigure}
     \begin{subfigure}[b]{0.4\textwidth}
         \centering
         \includegraphics[width=0.9\textwidth]{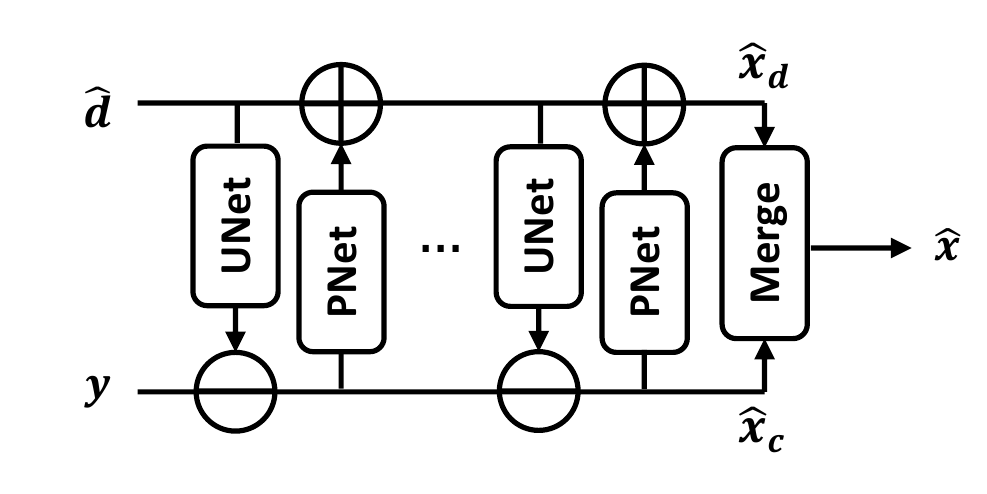}
          \vspace{-0.4cm}
         \caption{}
         \label{fig:three sin x}
     \end{subfigure}
        \caption{{\color{black}{(a) The forward part of 1-level WINN. (b) The backward part of 1-level WINN.}}}
        \label{fig:1level}
    \vspace{-0.3cm}
\end{figure}

\section{Proposed method}
In this paper, we propose an INN-guided probabilistic diffusion algorithm for general inverse problems. During the $T$-step diffusion process, we leverage INN to give $\bx_{t}$ a better update direction to ensure data consistency. In this section, we first introduce the design of the proposed wavelet-inspired invertible neural network (WINN), and then we describe how it works in the diffusion process.

\subsection{Wavelet-inspired Invertible Neural Network}
The framework of our proposed Wavelet-inspired Invertible Neural Network is shown in Fig.~\ref{fig:framework}. 
To preserve the perfect reconstruction property, we design our model by generalizing the prediction and update operators in the traditional lifting scheme \cite{, daubechies1998factoring} to deep networks called Predict and Update Networks (see PUNets in Fig.~\ref{fig:framework}). 
Different from the role INN played in \cite{winnet} as described in Sec. \ref{background_inn},
we propose to exploit the invertibility of INN and we treat it as a simulator of the degradation and a reconstruction process as follows.

\textbf{Simulating the Degradation Process:} Let us first assume our WINN has only one level. The forward transform of WINN can be represented as $\mathbf{c},\mathbf{d}= \textit{WINN}_{F}(\bx)$. As shown in Fig.~\ref{fig:1level}(a), the image is split into two parts by a splitting operator. Then the Predict Network (PNet) conditioned on the coarse part aims to predict the detail part, while the Update Network (UNet) conditioned on the detail part is used to adjust the coarse part to make it smoother. The Predict and Update networks are applied alternatively and finally generate the coarse and detail parts $\mathbf{c}$ and $\mathbf{d}$.

The Predict and Update networks can be any function and their properties will not affect the invertibility of WINN. In our implementation, one level of WINN is composed of Lazy Wavelet Transform and $M$ pairs of Predict/Update networks, as shown in Fig.~\ref{fig:1level}. Our Predict/Update network consists of an input convolutional layer, $J$ residual blocks with depth-wise separable convolution layers, and an output convolutional layer.

To model the degradation process,  we impose that $\mathbf{c}$ resembles $\mathbf{y}$. 
Given a training set $\left \{ \bx^{i}, \by^{i}\right \}_{i=1}^{N}$, which contains $N$ high-quality images and their low-quality counterparts, we optimize our WINN with the following loss function: 
\begin{align}
\begin{split}
L\left ( \Theta  \right )=\frac{1}{N}\sum_{i=1}^{N}\left \| \mathbf{c}^{i}-\by^{ i } \right \|_{2}^{2},
\end{split}
\end{align} 
where $\Theta$ denotes the learnable parameter set in our WINN.
Once we constrain one part of the output of $\textit{WINN}_{F}(\bx)$ to be close to $\by$, due to invertibility, the other part of the output $\mathbf{d}$ will inevitably represent the detailed information lost during the degradation process.

\textbf{Reconstrucion Process:} 
By construction, the inverse transform of WINN can perfectly recover the input original image from $\mathbf{c}$ and $\mathbf{d}$.
However, the goal of inverse problems is to reconstruct $\mathbf{x}$ from measurements $\mathbf{y}$ without the lost detail $\mathbf{d}$. Therefore, we propose to estimate the details ${\mathbf{d}}$ with a pre-trained diffusion model. As shown in Fig.~\ref{fig:1level}(b), the estimated detail $\hat{\mathbf{d}}$ and the measurement $\mathbf{y}$ will be transformed back to the reconstructed image $\hat{\mathbf{x}}$ using the inverse transform of WINN.
Thus, we improve the perceptual quality with the powerful generative prior provided by the diffusion model and ensure data consistency with INN.

\textbf{Multi-scale Property:}
Multi-resolution signal decomposition is an essential property of the wavelet transform, as shown in Fig.~\ref{fig:lifting}(b). And our WINN is able to generalise to multiple levels as well. While the split operator can be any invertible function that partitions the output space into two, we implement it with a non-redundant Lazy Wavelet Transform to partition the input based on sample parity. Therefore, we can simulate various sizes of $\mathbf{y}$ without constraining $\mathbf{y}$ and $\mathbf{x}$ to be of the same size. For example, if the degradation model contains a 2x/4x/8x downsampling operator, we can apply a 1/2/3-level WINN to solve the inverse problem.

\subsection{INN-Guided Probabilistic Diffusion Algorithm}
In the unconditionally trained DDPM \cite{ddpm}, the reverse diffusion process iteratively samples $\mathbf{x}_{t-1}$ from $p(\mathbf{x}_{t-1}|\mathbf{x}_{t})$ to yield clean images $\mathbf{x}_{0} \sim q(\mathbf{x})$ from initial random noise $\mathbf{x}_{T} \sim \mathcal{N}(\mathbf{0},\mathbf{I})$. 
{\color{black}{Here, we rewrite Eq.~\ref{eq:ddpm reverse 1} with the pre-trained approximator $\boldsymbol{\epsilon}_{\boldsymbol{\theta}}(\mathbf{x}_{t},t)$ and split it into the following two equations:
\begin{align}
\begin{split}
{\color{black}{\bx_{0,t}  = \frac{1}{\sqrt{\bar\alpha_t}}(\bx_{t} - \sqrt{1 - \bar\alpha_t} \bepsilon_\theta(\bx_t, t) )}}
\end{split}
\label{X0}
\end{align}
and 
\begin{align}
\begin{split}
{\color{black}\mathbf{x}_{t-1}=
{\frac{\sqrt{\alpha_t}(1-\bar\alpha_{t-1})}{1 - \bar\alpha_t}\bx_{t}+\frac{\sqrt{\bar\alpha_{t-1}}\beta_t}{1 - \bar\alpha_t}\bx_{0,t}  + \sigma_t \bz}}.
\end{split}
\end{align}
As illustrated in Eq. \ref{X0}, $\mathbf{x}_{0,t}$ is the predicted clean image from the noisy image $\mathbf{x}_{t}$.}} {\color{black}{To solve inverse problems, we need to refine each unconditional transition using $\by$ to ensure data consistency. }}In our proposed algorithm, we impose our data-consistency step by optimizing the clean image $\mathbf{x}_{0,t}$ instead of the noisy image $\mathbf{x}_{t}$.

As shown in Algorithm~\ref{1}, we impose an additional data consistency step after each original unconditional sampling update. 
At this additional step, we decompose the intermediate result $\mathbf{x}_{0,t}$ with WINN$_F$ into coarse $\mathbf{c}_{t}$ and detail part $\mathbf{d}_{t}$ and replace the coarse part $\mathbf{c}_{t}$ with the measurements $\by$. The INN-optimized $\hat\bx_{0,t}$ is generated by the inverse process WINN$_I$. Thus, the INN-optimized $\hat\bx_{0,t}$ is composed of the coarse information $\by$ and the details generated by the diffusion process. To incorporate the INN-optimized $\hat\bx_{0,t}$ into the DDPM algorithm, 
we update $\bx_{t}$ with the guidance of the gradient of $\|{\hat\bx_{0,t} - \bx_{0,t}}\|_2^2$. With the help of INN, our algorithm effectively estimates the details lost in the degradation process and is no longer limited by the requirement of knowing the closed-form expression of the degradation model.

\begin{minipage}[t]{0.46\textwidth}
    \vspace{-0.2cm}
\begin{algorithm}[H]
  \caption{DDPM Sampling with WINN} \label{alg:sampling}
  \small
  \begin{algorithmic}[1]
    \vspace{.04in}
    \State $\bx_T \sim \mathcal{N}(\bzero, \bI)$
    \For{$t=T, \dotsc, 1$}
      \State $\bz \sim \mathcal{N}(\bzero, \bI)$ if $t > 1$, else $\bz = \bzero$
      \State{{${\color{black}{\bx_{0,t}  = \frac{1}{\sqrt{\bar\alpha_t}}(\bx_{t} - \sqrt{1 - \bar\alpha_t} \bepsilon_\theta(\bx_t, t) )}}$}}
        \State{$\tilde{\bx}_{t-1}  = \frac{\sqrt{\alpha_t}(1-\bar\alpha_{t-1})}{1 - \bar\alpha_t}\bx_{t}+\frac{\sqrt{\bar\alpha_{t-1}}\beta_t}{1 - \bar\alpha_t}\bx_{0,t}  + \sigma_t \bz$}
              \State {\color{blue}$\mathbf{c}_{t},\mathbf{d}_{t}= \textit{WINN}_{F}(\bx_{0,t})$}
            \State {\color{blue}$\hat\bx_{0,t}=\textit{WINN}_{I}(\by,\mathbf{d}_{t})$}  
                \State{\color{blue}{$\bx_{t-1} =\tilde{\bx}_{t-1}  - { {\zeta}}\nabla_{\bx_{t}} \|{\hat\bx_{0,t} - \bx_{0,t}}\|_2^2$}}
    \EndFor
    \State \textbf{return} $\bx_0$
    \vspace{.04in}
  \end{algorithmic}
  \label{1}
\end{algorithm}
\end{minipage}

\section{Experiments}

\subsection{Implementation Details}
Empirically, we set $\zeta$=0.5, $M$=4, $J$=2 and $T$=1000. The number of feature channels in PUNet is set to 32 and the spatial filter size in depth-wise separable convolutional layers is set to 5. The total number of learnable parameters of our WINN is 0.71M. 

We test our method on the FFHQ 256×256 1k validation dataset \cite{ffhq}, CelabA HQ 256×256 1k validation dataset \cite{celeba}, and a real-world SR dataset DRealSR \cite{drealsr}. For the face photo reconstruction, we utilize a pre-trained unconditional diffusion model trained on the FFHQ training dataset by \cite{dps} and select 1k images from the FFHQ training dataset to train our WINN.
For the natural image reconstruction on \cite{drealsr}, we utilize the pre-trained unconditional diffusion model trained on ImageNet \cite{Imagenet} by \cite{dhariwal2021diffusion} and use DRealSR \cite{drealsr} training dataset to train our WINN. 
We apply our proposed method to three settings for inverse problems:  downsampling with/without noise, complex degradation model based on combining downsampling with jpeg compression, and real-world degradation model. The reconstruction results are evaluated with PSNR, FID \cite{fid}, LPIPS \cite{lpips} and NIQE \cite{niqe}.

\begin{table}[t]
\centering
\vspace{-5pt}
\caption{Quantitative results on the problems of 4× super-resolution with noise on different levels on the FFHQ 1k validation dataset. The best results are highlighted.}
\setlength{\tabcolsep}{8pt}  
\vspace{2pt}
\label{tab:bicubic}
\begin{tabular}{cccccc}
\hline
 Method & Noise $\sigma$& PSNR  $\uparrow$   & FID $\downarrow$ & LPIPS $\downarrow$&NIQE$\downarrow$ \\ \hline
 ILVR  &0 & 27.43&44.04 &0.2123 &5.4689 \\
 DDRM  &0 &{28.08} &65.80 &0.1722& 4.4694\\
DPS& 0&26.67 &32.44 & 0.1370 &4.4890   \\
Ours &0& \textbf{28.15}     &\textbf{22.33} & \textbf{0.0889 }& \textbf{4.1564} \\ \hline
   ILVR  &0.05 &26.42 & 60.27&0.3045 & 4.6527\\
 DDRM  & 0.05&{27.06} &45.90  &0.2028& 4.8238\\
DPS &0.05 &25.92&31.71 &0.1475&4.3743 \\
Ours  & 0.05& \textbf{27.16}& \textbf{26.64}&\textbf{0.1215}&\textbf{4.1004} \\ \hline
 ILVR  &0.10 &24.60 &88.88 &0.4833 &4.4888 \\
 DDRM  & 0.10 & {26.16} & 45.49 & 0.2273 & 4.9644\\
DPS & 0.10 & 24.73 &31.66 &0.1698&4.2388 \\
Ours  & 0.10& \textbf{26.25} &\textbf{28.89} &\textbf{0.1399}& \textbf{3.9659}\\ \hline

\end{tabular}
\end{table}

\subsection{Results with Noiseless and Noisy Downsampling Degradation Model}
\label{sr}
\begin{figure}[t]
    \centering
    \includegraphics[width=0.48\textwidth]{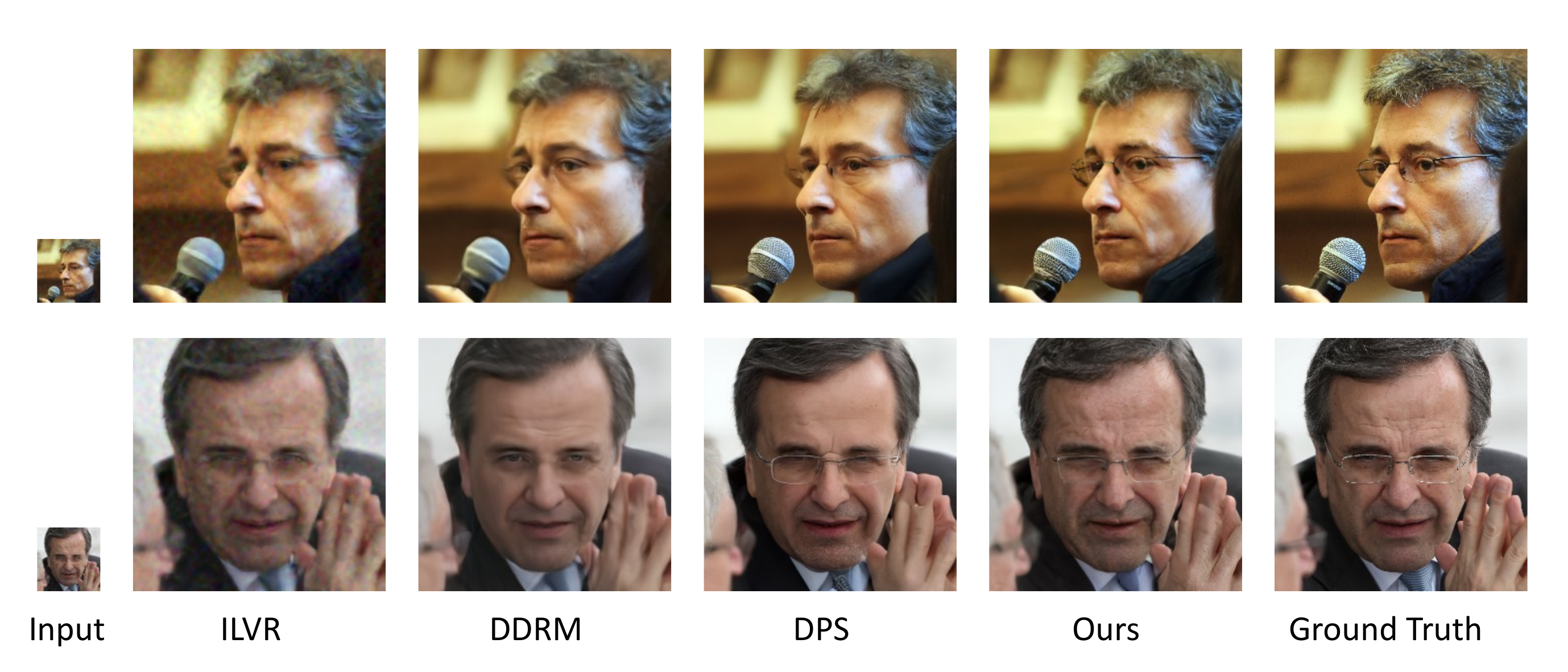}
\caption{Visual comparisons on solving the super-resolution problem (x4) with $\sigma_y $= 0.05 on FFHQ validation dataset.}
    \label{fig:com}
\end{figure}

We compare our method with 3 state-of-the-art methods based on diffusion models: ILVR \cite{ilvr}, DDRM \cite{DDRM}, and DPS \cite{dps}. 
We evaluate all methods on the problems of 4× super-resolution with different levels of noise on the FFHQ dataset in {\color{black}{Table \ref{tab:bicubic}. One can observe that our method outperforms all baseline methods in all metrics, which means the distance between real and generated image distributions in our results is lower and the perceptual quality of our generated images is better. As can be seen in Fig. \ref{fig:com}, our algorithm produces high-quality reconstruction results and preserve more details than other methods. }}

\begin{figure}[t]
    \centering
    \includegraphics[width=0.48\textwidth]{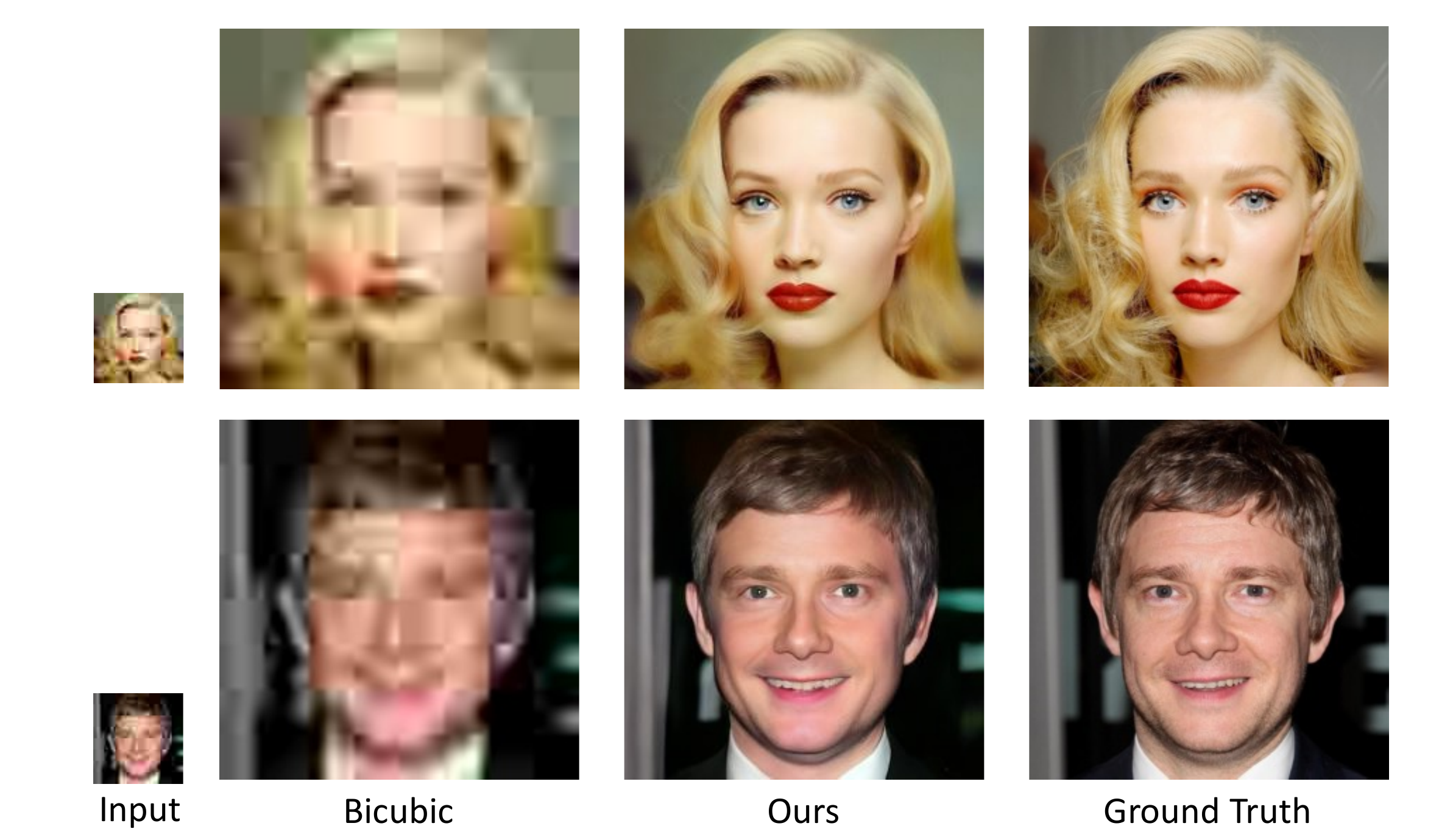}
\caption{Results of our algorithm on solving the inverse problem in Eq.~\ref{jpeg} on CelabA HQ validation dataset.}
    \label{fig:jpeg10}
                \vspace{0.2cm}
\end{figure}

\subsection{Results with Complex Degradation Model}
\label{complex degradation model}

As described in the introduction, our flexible WINN can handle a variety of degradation processes. Here we take the following degradation process as an example to evaluate the performance with a non-linear, complex degradation model:
\begin{align}
    \mathbf{y}& =JPEG_{10}(\mathbf{x}\downarrow_4),
\label{jpeg}
\end{align}
where the jpeg factor and downsampling scale are 10 and 4, respectively.
As shown in Fig. \ref{fig:jpeg10}, our algorithm can still produce high-quality images from heavily degraded input measurements. It can be seen that our results produce realistic details while ensuring data consistency. Note that the methods we compared against in Sec. \ref{sr} are not able to support this type of degradation, so for this case, we only show our results.

\begin{figure}[t]
    \centering
    \includegraphics[width=0.45\textwidth]{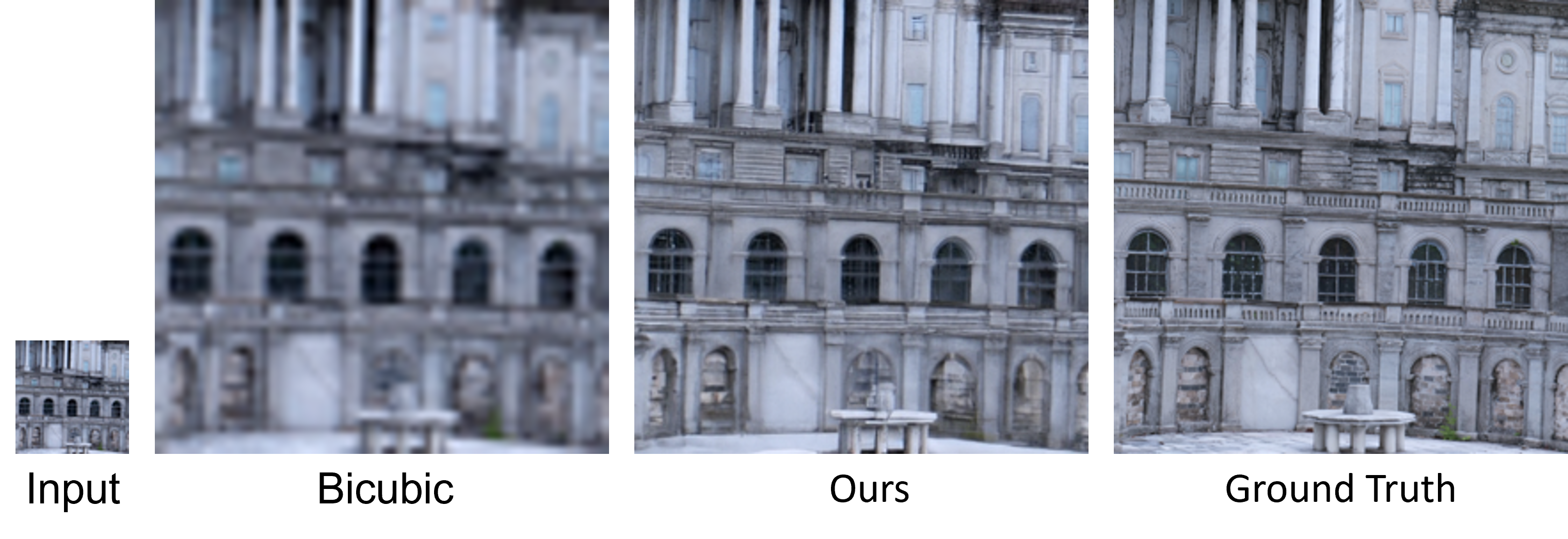}
\caption{ {\color{black}{Result of our algorithm on reconstructing real images from DRealSR \cite{drealsr} with resolution enhancement by a factor 4 per direction. We can observe that our algorithm can produce high-quality results for real-world degradation.}}}
    \label{fig:real}
\end{figure}

\begin{figure}[t]
    \centering
    \includegraphics[width=0.47\textwidth]{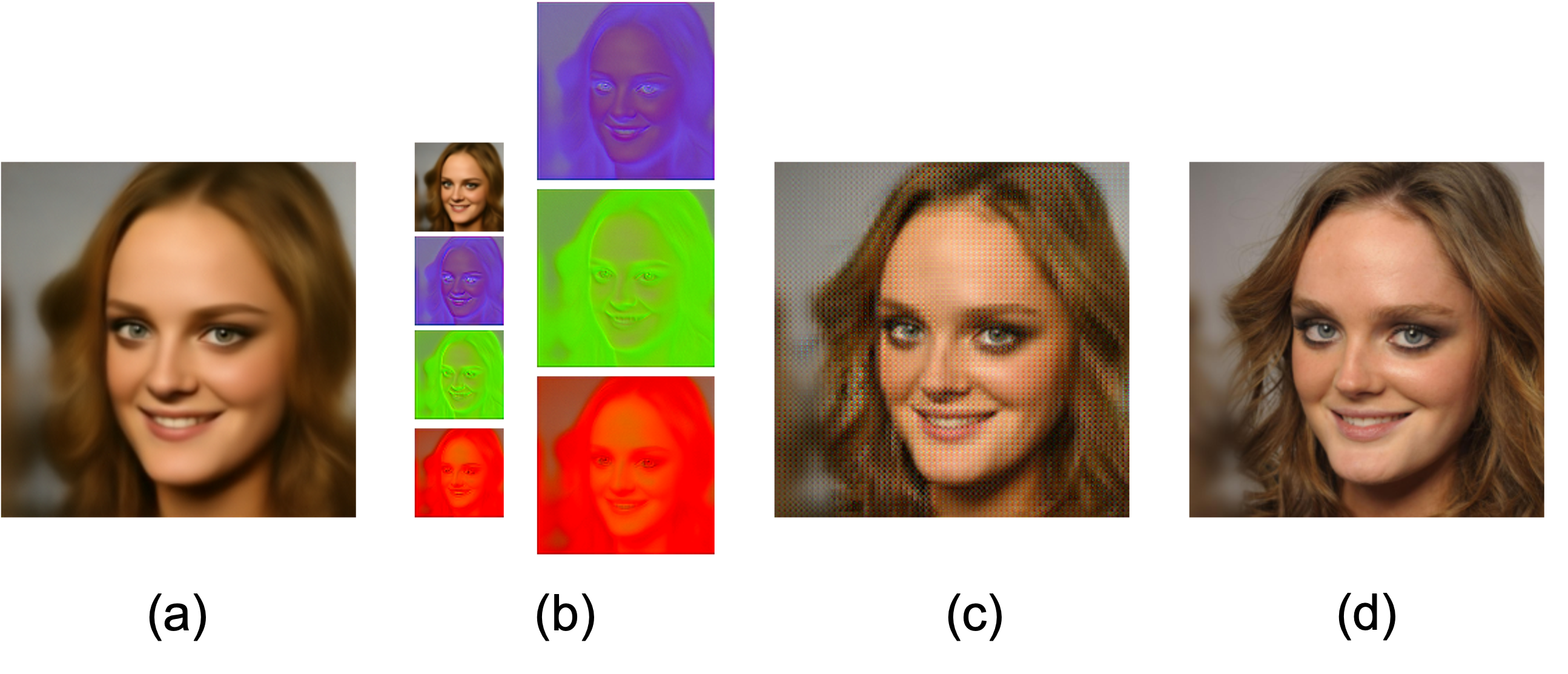}
\caption{Intermediate results in our INDigo. (a)  Input to the forward process WINN$_F$ at time step 500. (b) Output of the forward process WINN$_F$ at time step 500. (c) Output of the inverse process WINN$_I$ at time step 500. (d) Ground truth. }
    \label{fig:inninput}
\end{figure}

\subsection{Results with Real Degradation Model}
Finally, in Fig. \ref{fig:real}, we show the results of our algorithm on reconstructing images from real-world degradation processes using DRealSR \cite{drealsr} with scale factor 4.
As for Sec. \ref{complex degradation model}, since we simulate the degradation with INN, our solution is no longer limited by the requirement of knowing the closed-form expression of the degradation process. 
We can observe that our algorithm can produce, also in this case, high-quality results for real-world degradation.

\subsection{Analysis}
In order to further verify the role of INN in our algorithm, we show some intermediate results here.
Fig. \ref{fig:inninput}(a) and Fig. \ref{fig:inninput}(b) show the input and output of the forward process of WINN (see the 6-th line of Algorithm~\ref{1}) and Fig. \ref{fig:inninput}(c) shows the output of the reverse process of WINN (see the 7-th line of Algorithm~\ref{1}). These results effectively verify the update direction given by INN to ensure data consistency.

\section{Conclusion}
In this paper, we have proposed an INN-Guided Probabilistic Diffusion Algorithm (INDigo) for general inverse problems, which combines the merits of the perfect reconstruction property of INN and the strong generative capabilities of diffusion models. During the diffusion process, we ensure data consistency by imposing an additional step with our proposed Wavelet-based Invertible Neural Network (WINN). By introducing INN in diffusion posterior sampling, our algorithm effectively estimates the details lost in the degradation process and does not require the closed-form expression of the degradation model. Experiments demonstrate that our algorithm achieves state-of-the-art results compared with recently leading methods and performs well on more complex degradation models and on real-world low-quality images.

\bibliographystyle{IEEEtran}
\small
\bibliography{conference_101719}

\vspace{8pt}

\end{document}